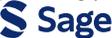

# CitySim: A Drone-Based Vehicle Trajectory Dataset for Safety-Oriented Research and Digital Twins

Ou Zheng[1] [iD], Mohamed Abdel-Aty[1] [iD], Lishengsa Yue[1,2] [iD], Amr Abdelraouf[1] [iD], Zijin Wang[1] [iD], and Nada Mahmoud[1] [iD]

## Abstract

The development of safety-oriented research and applications requires fine-grain vehicle trajectories that not only have high accuracy, but also capture substantial safety-critical events. However, it would be challenging to satisfy both these requirements using the available vehicle trajectory datasets do not have the capacity to satisfy both. This paper introduces the CitySim dataset that has the core objective of facilitating safety-oriented research and applications. CitySim has vehicle trajectories extracted from 1,140-min of drone videos recorded at 12 locations. It covers a variety of road geometries including freeway basic segments, weaving segments, expressway merge/diverge segments, signalized intersections, stop-controlled intersections, and control-free intersections. CitySim was generated through a five-step procedure that ensured trajectory accuracy. The five-step procedure included video stabilization, object filtering, multivideo stitching, object detection and tracking, and enhanced error filtering. Furthermore, CitySim provides the rotated bounding box information of a vehicle, which was demonstrated to improve safety evaluations. Compared with other video-based trajectory datasets, CitySim had significantly more safety-critical events, including cut-in, merge, and diverge events, which were validated by distributions of both minimum time-to-collision and minimum post encroachment time. In addition, CitySim had the capability to facilitate digital-twin-related research by providing relevant assets, such as the recording locations' three-dimensional base maps and signal timings.

Traffic safety research has rapidly progressed over the past few years. Example applications include autonomous vehicle safety, proactive traffic safety, and connected traffic safety applications. This increasing interest in safety research has stimulated a rising demand for vehicle trajectory datasets. Vehicle trajectory datasets can be collated by onboard sensors, roadside cameras, radars, and several other types of data collection equipment. Among these datasets, birds-eye-view video-based trajectory datasets have been attracting increasing attention. Currently, a significant portion of related studies have been formulated using video-based trajectory datasets. These studies include driver behavior analysis (1–3), autonomous vehicle virtual testing (4–6), crash mitigation and avoidance system design (7–10), advanced autonomous control algorithm development (11–14), surrogate safety measures (15–18), among other safety applications. In the autonomous vehicle safety field particularly, datasets can provide training/testing samples for personalized algorithms of trajectory/intention prediction and vehicle control, given that the dataset contains rich information about driving preferences.

The major advantages of extracting road user trajectories from birds-eye-view videos are 1) the extracted

[1]Department of Civil, Environmental & Construction Engineering, University of Central Florida, Orlando, FL
[2]The Key Laboratory of Road and Traffic Engineering, Ministry of Education, Tongji University, Shanghai, China

**Corresponding Authors:**
Ou Zheng, ouzheng1993@knights.ucf.edu
Lishengsa Yue, 2017lishengsa@Knights.ucf.edu

trajectories are comprehensive, 2) the trajectories are free from occlusion-related problems, and 3) vehicle geometries and intervehicle distances are easily preserved. Other trajectory extraction methods include data collection vehicles and roadside cameras. Data collection vehicles can accurately capture nearby vehicle trajectories using onboard sensors; however, more distant objects are easily missed owing to sensor range limitations and object occlusion. Roadside cameras capture road users at an angle, thus, the actual vehicle geometries and intervehicle distances are distorted. In contrast, birds-eye-view videos can cover most trajectories within a specified detection area while preserving a vehicle's shape, which are essential for safety-critical applications.

Previous research efforts that utilized birds-eye-view video-based trajectories, particularly in the traffic safety domain, have outlined some of the limitations of the available datasets. For instance, existing birds-eye-view video-based trajectory datasets were not specifically designed for safety research. Therefore, they tend to lack a significant number of safety-critical events. The most well-known and widely used video-based trajectory datasets are NGSIM (*19*), HighD (*20*), InD (*21*), and Interaction (*22*). NGSIM was proposed in 2002; however, because of technological limitations in the early years, the dataset contains several trajectory errors (*23–25*). The dataset has a reported false negative issue that can cause more than 10% of the vehicle detection and tracking process to fail (*26*). Consequently, to remedy this issue, previous research has usually conducted filtering processes before using the dataset (*25*). Furthermore, NGSIM has very few near-collision events (*22*).

HighD, InD, and Interaction were introduced between 2018 and 2019. They use more advanced computer vision algorithms to accurately extract vehicle trajectories. HighD mainly contains freeway segments under free-flow conditions, in which intensive vehicle interactions and safety-critical events are inevitably limited (*27*). Consequently, safety research might not be able to collect sufficient valuable samples from HighD. InD and Interaction datasets were extracted from intersections. InD is limited to four, small, nonsignalized intersections, whereas Interaction mainly comprises nonsignalized intersections and roundabouts. Another major issue with these aforementioned datasets is that the rotated bounding box information of a vehicle is not provided. As demonstrated in later sections of this paper, rotated bounding box information is essential for robust safety-related measurements and analysis.

Recently, safety researchers have started to investigate digital-twin applications that leverage advanced virtual simulation. A digital twin, in transportation, is a digital replica of all the traffic participants. Kumar et al. used a digital-twin-based technology to predict driver intention and improve traffic operation (*28*). Wang et al. designed a digital-twin-assisted cooperative strategy for nonsignalized intersections (*29*). Schwarz and Wang discussed the role of digital twins in connected and automated vehicles, and they predicted the future paradigm of the digital-twin services (*30*). By providing a virtual environment that is an exact replica of the real world, autonomous vehicles can be tested more thoroughly with virtual onboard sensors. However, existing birds-eye-view video-based trajectory datasets do not encompass features that support digital-twin concepts.

In this paper, we introduce the CitySim dataset: a birds-eye-view video-based trajectory dataset generated from drone recordings with a focus on traffic safety research and applications. It was designed with the following novel features:

1) Accurate vehicle trajectories. CitySim has a five-step procedure to ensure high trajectory accuracy: video stabilization, object filtering, multivideo stitching, detection and tracking using an integrated algorithm of a Mask region-based convolutional neural network (Mask R-CNN [*31*]) and a channel and spatial reliability tracker (CSRT [*32*]), and enhanced error filtering. Algorithms and data processing methods are further delineated to give the dataset users a clear idea about the quality of the extracted trajectories.

2) Wider vehicle trajectory range. Because of drone flight altitude restrictions and video resolution limitations, the coverage of drone cameras is limited. To capture wider observation areas, CitySim utilizes multiple drones to hover over target areas and subsequently stitches the generated videos into one cohesive video. Therefore, CitySim users can observe and analyze traffic safety events from a much longer trajectory range, which accounts for various scenarios and provides comprehensive results.

3) More safety-critical events. CitySim selects locations that contain more aggressive and intensive vehicle interactions, such as weaving segments. As demonstrated later in this paper, CitySim has more safety-critical events compared with other datasets in relation to both safety severity level and sample size.

4) Accurate vehicle geometric representation. CitySim provides rotated bounding box information to represent the vehicle geometries for each detected vehicle, which enables a more accurate estimation of surrogate safety measures. This paper demonstrates the necessity of using rotated bounding box information to calculate surrogate

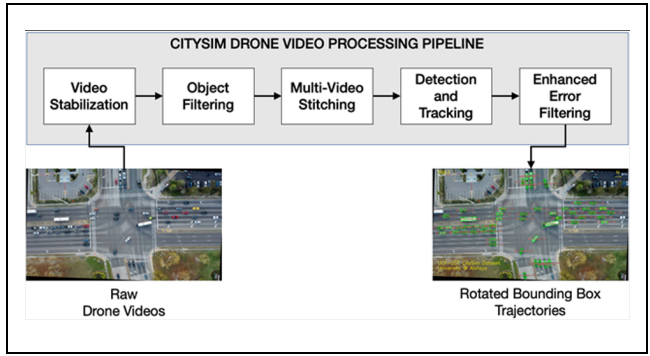

**Figure 1.** Production pipeline of CitySim.

safety measures, compared with only using the vehicle's central point information.

5) Digital-twin features. CitySim provides high resolution three-dimensional (3D) maps and physical models for each of the selected locations. It additionally provides signal timing information. These assets allow researchers to test and verify their safety research in a digital-twin virtual environment.

CitySim has been supporting the many research efforts of its development team, the University of Central Florida Smart and Safe Transportation (UCF-SST) Lab. Examples include crash risk evaluations (33, 34), trajectory predictions (35), and computer-vision-based surrogate safety measurement (35, 36). Moreover, since the release of its beta version in April 2022, CitySim has been attracting attention from researchers and institutions worldwide. As of August 1, 2022, over 50 research teams from the United States, China, Europe, the Middle East, and many other places have submitted applications to use CitySim.

The remainder of this paper is organized as follows: the next section explains the processing pipeline used to generate CitySim, including key details of its algorithms. The subsequent section describes CitySim's locations and attributes. Following this, a statistical comparison of safety-critical events between CitySim and other datasets is presented. The digital-twin assets and features are then described. The final section summarizes our conclusions and sets out suggestions for future work.

## Dataset Generation

CitySim was generated through the five steps depicted in Figure 1. The five steps include video stabilization, object filtering, multivideo stitching, detection and tracking, and enhanced error filtering. These were put in place to ensure that the output trajectories are as accurate as possible.

### Video Stabilization

Since drone stability may be affected by unsteady airflow and vibration, the output video must be stabilized to obtain reliable trajectories. First, the scale-invariant feature transform (SIFT) algorithm (37) was used to extract image features from the first and last frames of the video, to collect shared features across the frames. Next, random frames were sampled and stabilized using the extracted SIFT features. Afterwards, the median pixels from the selected frames were calculated to build a vehicle-free accumulated weighted frame. Each video frame was then mapped to the accumulated weighted frame using a homography transformation. Homography, also known as planar homography, is a transformation that takes place between two planes (38). In other words, it maps between two image planar projections. In a homogeneous coordinate space, this is represented by a $3 \times 3$ transformation matrix.

When a drone is at high altitude, small movements, such as from swaying vegetation, can cause SIFT mapping failures. SIFT creates a keypoint descriptor with $16 \times 16$ neighbors around the keypoint. Keypoints between two images are matched by identifying their nearest $16 \times 16$ neighbors. Sometimes, owing to noise or because the object itself is moving for example, the second-closest match may be very near the first. SIFT applies a ratio of the closest distance to the second-closest distance to handle such cases. If the distance is more significant than 0.8, the keypoints are rejected. SIFT eliminates around 90% of false matches, while discarding only 5% of correct matches. However, the video stabilization function can sometimes still fail.

To solve this failure mapping issue, a CSRT was used as a backup to match features across the frames. Figure 2 depicts an example of video stabilization.

### Object Filtering

The video recordings contain some road markings or other objects that may cause detection errors, particularly false positives. Therefore, these objects must be eliminated. First, vehicles were removed from the background using a foreground/background segmentation Gaussian-mixture-based algorithm (39). Then the inpainting algorithm (40) was used to remove any remaining undesirable objects in the background. Removing road markings reduces the number of vehicle detection false positives. Figure 3 depicts the object filtering operation.

### Multivideo Stitching

Owing to flight height restrictions and video resolution limitations, a single drone can only cover a limited observation area. Therefore, to increase the length of the recorded trajectories in CitySim, a multivideo stitching process was used to combine videos from multiple drones. Each pair of videos was stitched based on histogram color matching, SIFT features, and image blurring (*41*). The histogram color matching algorithm uses the color distribution to unify color brightness and contrast across the video pairs. This step is necessary to avoid reidentification errors when vehicles cross from one drone image to another. Next, SIFT features were extracted from the overlapping area in each video; each matching feature pair was used for stitching. Figure 4 depicts the multivideo stitching process.

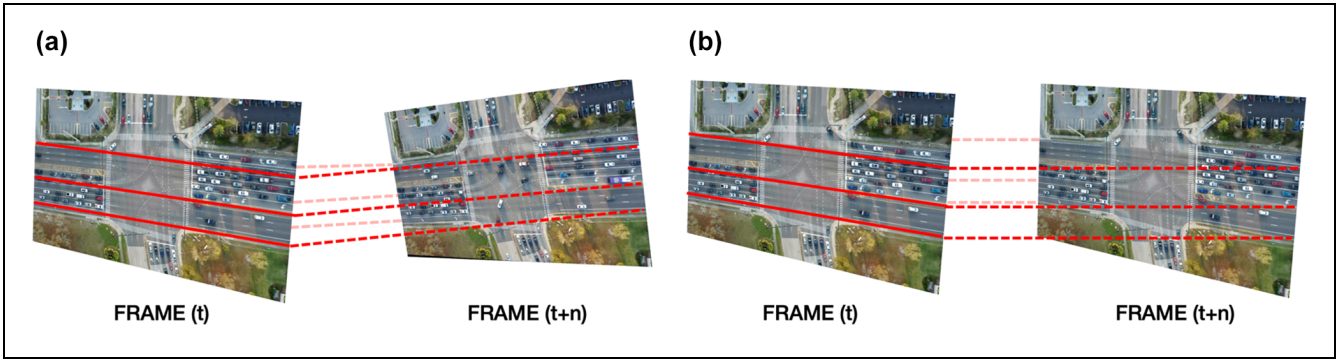

**Figure 2.** Video stabilization: (*a*) before and (*b*) after.

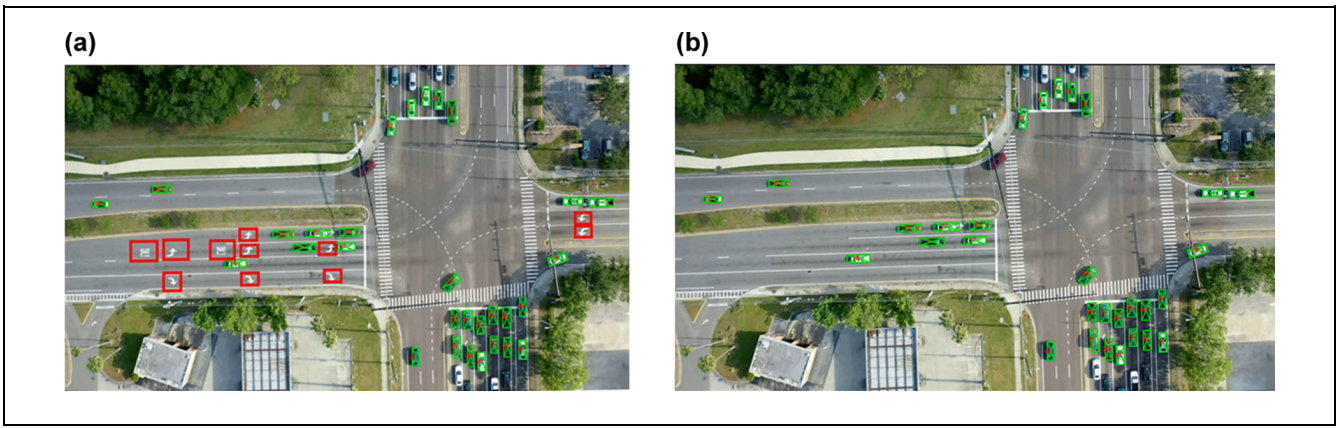

**Figure 3.** Object filtering from the video recording: (*a*) before and (*b*) after.

### Object Detection and Tracking

Object detection and tracking algorithms can be divided into one-stage algorithms (e.g., YOLO and its follow-up improvements [*42–44*]), multistage algorithms (e.g., R-CNN [*45*], fast R-CNN [*46*]), and anchor free algorithms (*47*, *48*). In this study, Mask R-CNN was used for vehicle detection in the video recordings. For each detected vehicle, the algorithm generated a segmentation mask in a pixel-to-pixel manner (*31*). Based on the masks, a vehicle's bounding box was rotated according to the direction of movement. The idea is to rotate a spring-loaded vernier caliper around the outside of a Mask R-CNN output mask. This can be thought of as a polygon. When one of the caliper's blades lies flat against the polygon's edge, it forms an antipodal pair with the point or edge touching the opposite blade. The caliper rotation detects all antipodal pairs. The minimum rectangle of this polygon is then formed by viewing this set of all pairs as a graph.

Most traditional video detection algorithms generate straight bounding boxes that do not align with the vehicle's direction of movement. Thus, the resulting vehicle sizes tend to be larger when the vehicles are on a curve or turning at an intersection. Figure 5 illustrates the differences between the detected vehicle's mask, a straight bounding box, and a rotated bounding box. The rotated bounding box has been demonstrated to more accurately represent the detected vehicle size, orientation, and location. Moreover, CSRT was used for tracking the objects that were detected by Mask R-CNN. In the case of a missed detection by Mask R-CNN, CSRT can provide an interpolated tracking location for the vehicle in question.

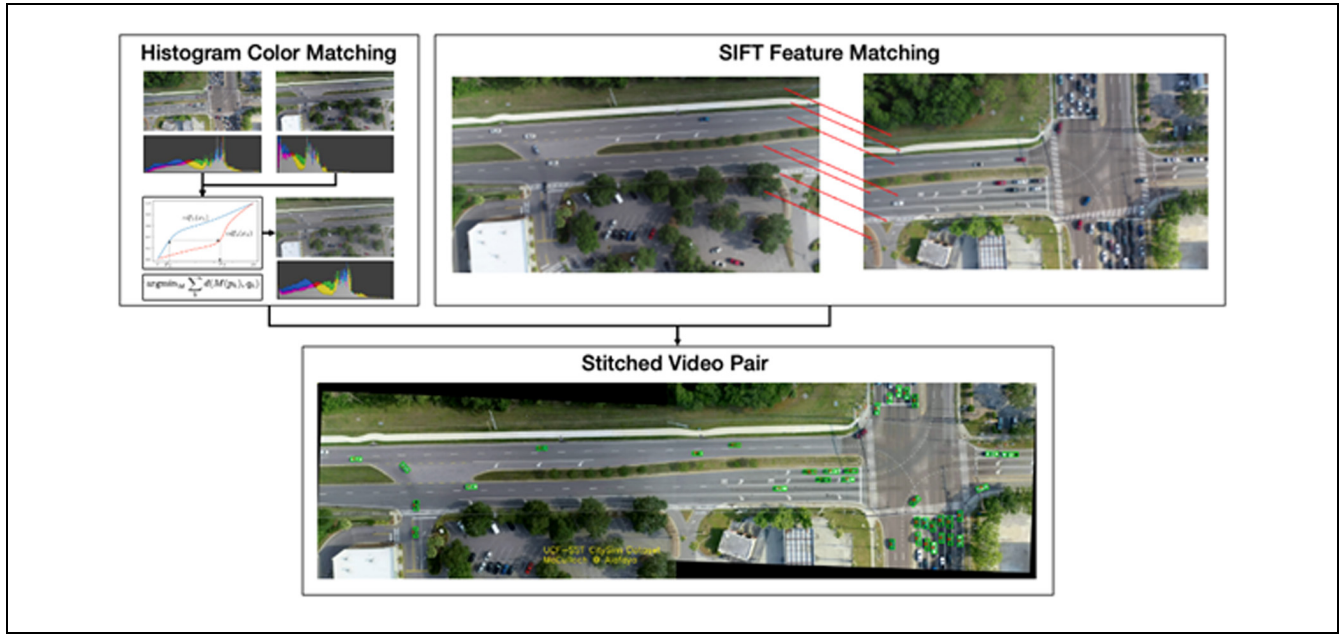

**Figure 4.** Drone-based multivideo stitching.

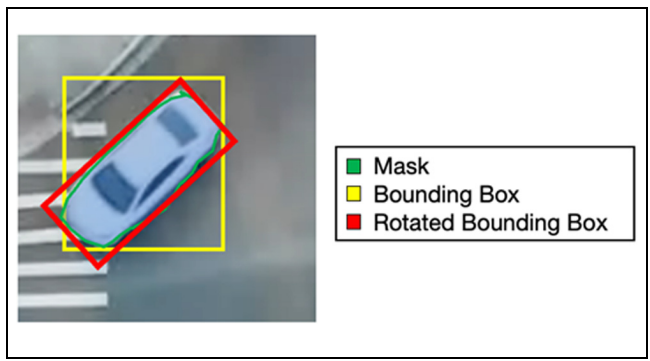

**Figure 5.** Straight bounding box versus rotated bounding box.

## Enhanced Error Filtering

The extracted bounding boxes may still have some errors because of environmental factors such as shadows from trees or buildings, inevitable detection errors, and unobserved vehicle types in the training dataset. Therefore, the UCF-SST team developed a data fixing tool to check for potential errors. The data fixing tool allows users to manually delete wrongly detected objects and adjust the sizes and headings of erroneous bounding boxes. Next, following the active learning paradigm (*49*, *50*), the adjusted data were returned to the training dataset and utilized in successive rounds of vehicle detection training to increase detection accuracy. Some 3% to 4% of the data were fixed to mitigate for factors such as the sun, shadows, and so forth. Figure 6 illustrates an example of adjusting an erroneous vehicle bounding box using the data fixing tool.

## Model Accuracy Evaluation

The accuracy of the proposed algorithm was evaluated based on intersection of union (IOU) values. IOU is calculated using the following equation:

$$IOU = \frac{DetectionResult \ \cap \ GroundTruth}{DetectionResult \ \cup \ GroundTruth} \quad (1)$$

Higher IOU values indicate higher accuracy of the detection and tracking results. The study adopted several methods to improve the data quality; in addition to the vehicle detection algorithm (i.e., Mask R-CNN detection with CSRT), this study also improved model accuracy by using a data fix tool to correct any deep learning model errors. Table 1 shows the change in accuracy from the initial condition to the final output. Twenty vehicles were randomly selected from the UAV (unmanned aerial vehicle) video. IOU values were calculated for the movement duration for each vehicle based on the outputs and ground truths that were collected manually. In total, 3,541 video images were collected to calculate the IOU for the selected vehicles. It can be seen from the data that the proposed algorithm significantly improved the performance for all 16 types of movement, but especially for turning movements (i.e., left and right turning). The straight movements had the best performance for both methods.

## Citysim Dataset Description

CitySim currently contains vehicle trajectories from 1,140 min of drone video recordings of 12 locations. These

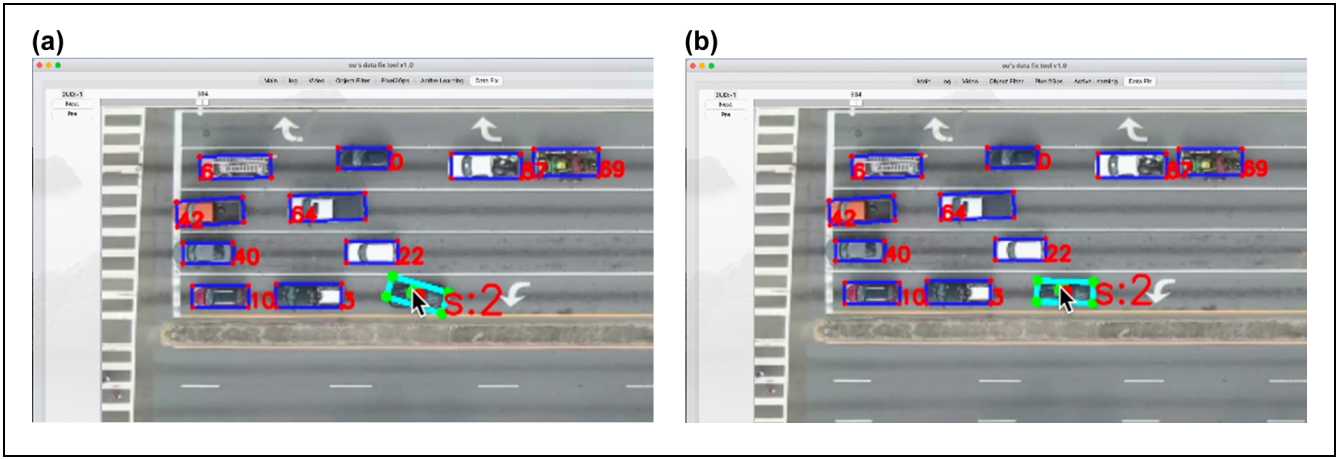

**Figure 6.** Using the data fixing tool to adjust bounding box errors (the light blue bounding box is fixed in the figure): (*a*) before error filtering and (*b*) after error filtering.

**Table 1.** IOU of the Proposed Algorithm and the Final IOU after Data Fix Tool

| Type of movement | Mask R-CNN detection and CSRT without rotation | Proposed algorithm | After data fix tool |
|---|---|---|---|
| Left turn | 0.42 | 0.78 | 0.98 |
| Right turn | 0.38 | 0.70 | 0.97 |
| Straight | 0.80 | 0.85 | 0.98 |
| Average | 0.53 | 0.76 | 0.976 |

*Note:* IOU = intersection of union; Mask R-CNN = Mask region-based convolutional neural network; CSRT = channel and spatial reliability tracker.

locations were selected because they have intensive vehicle interactions and conflicts. Vehicle trajectories contain valuable information about safety-critical events, and are therefore useful for safety-oriented research. Certain locations were selected to also provide a full picture of roadway facility categories.

The drone hardware models utilized were the DJI Mavic 2 and -3, and the Phantom 4. The locations in CitySim include one freeway basic segment, one weaving segment, two merge/diverge segments, five signalized intersections, one stop-controlled intersection, and two control-free intersections. The weaving segment and merge/diverge segments were specifically selected because they typically have a high number of safety-critical events, such as cut-in, merge, and diverge events. Details of the data collection methodology can be found in Table 2, and a birds-eye view of each of the 12 locations is illustrated in Figure 7.

The dataset trajectories are provided as comma separated value (CSV) files. Each row represents a waypoint (i.e., an intermediate point on a route or line of travel) that belongs to a vehicle trajectory in a single frame. Each waypoint contains the position information of seven vehicle key points: center point, head, tail, and four bounding box vertices, as depicted in Figure 8. Each position is provided in pixels, feet, and GPS coordinates. In addition, the speed, heading (measured in reference to both Global North and the image x-axis), and vehicle lane number are provided.

Figure 9 shows the space-time trajectory map at two typical locations, that is, a weaving segment (Expressway A) and an intersection (Intersection A). The trajectory map at the weaving segment could indicate interference from vehicle interactions. Vehicle speeds in the area delineated by an oval slowed because there were vehicles that wanted to leave the freeway overtaking the leading vehicles that also wanted to enter the off-ramp. Such vehicle interaction interrupts the traffic flow. The accumulated trajectory map at the intersection. A shows the traffic volume at the different intersection approaches. It can be seen that the volume of turning traffic was sizable at this location.

## Toward Safety Research

CitySim was designed to facilitate safety research such as autonomous vehicle safety and surrogate safety measures. In this section, CitySim is compared with two widely used trajectory datasets, that is, NGSIM and HighD, in relation to the following potential safety events:

- Freeway cut-in event: A cut-in vehicle may have a significant conflict with the following vehicle on the target lane. In this study, a cut-in event exclusively referred to an event occurring on the mainline, without any intention to merge or diverge.

**Table 2.** Data Collection Information

| ID | Location | Location type | Drone height (m)[b] | FPS | Recording resolution (pixel) | Recording length (min) |
|---|---|---|---|---|---|---|
| 1 | Expressway A | Weaving segment | 120 | 30 | 5120 × 2880 | 120 |
| 2 | Freeway B | Basic segment | 320 | 30 | 5120 × 2880 | 60 |
| 3 | Freeway C | Merge/diverge segment | 320 | 30 | 5120 × 2880 | 60 |
| 4 | Expressway D | Merge/diverge segment | 120 | 30 | 3840 × 2160 | 60 |
| 5 | Intersection A | Signalized intersection | 120 | 30 | 3840 × 2160 | 120 |
| 6 | Intersection B | Signalized intersection | 120 | 30 | 3840 × 2160 | 120 |
| 7 | Intersection C | Signalized intersection | 120 | 30 | 4096 × 2160 | 120 |
| 8 | Intersection D | Signalized intersection | 120 | 30 | 3840 × 2160 | 120 |
| 9 | Intersection E | Signalized intersection | 120 | 30 | 3840 × 2160 | 120 |
| 10 | Intersection F | Stop-control intersection | 120 | 30 | 3840 × 2160 | 120 |
| 11 | Intersection G | Control-free intersection[a] | 120 | 30 | 3840 × 2160 | 60 |
| 12 | Intersection H | Control-free intersection | 120 | 30 | 3840 × 2160 | 60 |

[a]Control-free intersection is neither signalized nor stop-controlled.
[b]Drone height is calculated from the take-off surface.

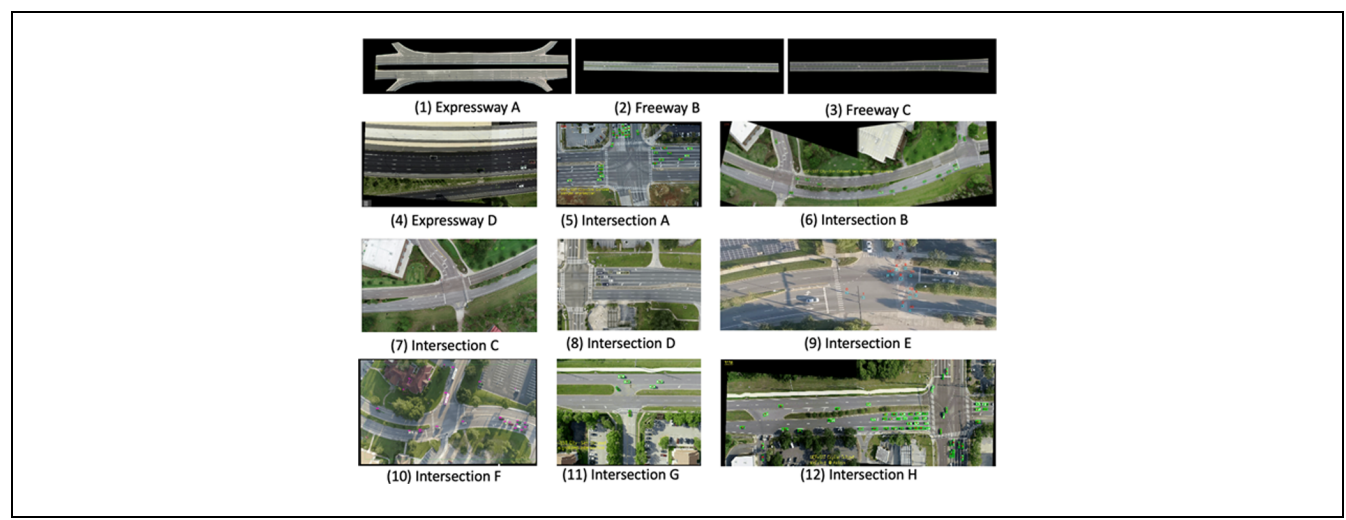

**Figure 7.** Birds-eye view of the CitySim locations.

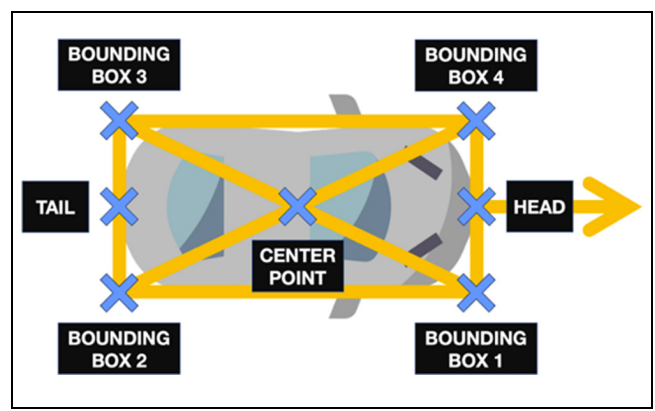

**Figure 8.** Vehicle bounding box feature description.

The cut-in is one type of safety-critical event that may cause a rear-end crash. Moreover, it has been observed that when a vehicle begins a lane change, the vehicle that was following it in the original lane often accelerates and thus generates another potential conflict. For a cut-in event, two types of minimum time-to-collision (minTTC) *(51)* metrics during the event were computed: a minTTC before cut-in, and a minTTC after cut-in. The before-minTTC is the minTTC between the cutting-in vehicle and the following vehicle in the original lane, whereas the after-minTTC is the minTTC between the cutting-in vehicle and the following vehicle in the target lane. Figure 10 depicts the two conflict types.

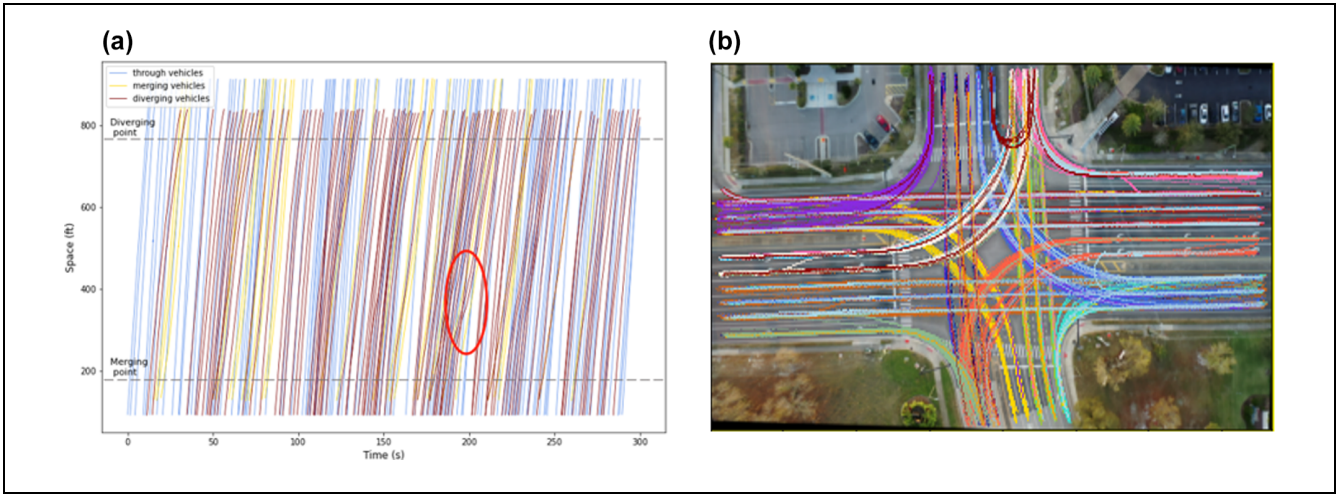

**Figure 9.** Vehicle trajectory map at Expressway A and Intersection A: (*a*) space-time trajectory at Expressway A and (*b*) accumulated trajectory at Intersection A.

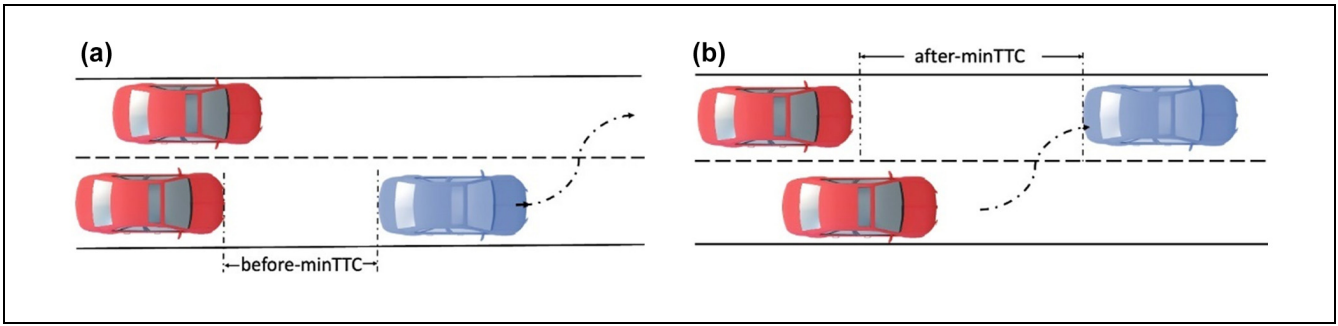

**Figure 10.** Two types of minimum time-to-collision (minTTC) during a cut-in event (*a*) before cut-in and (*b*) after cut-in.

- Freeway merge/diverge event: In this study, a merge event is defined as a vehicle from a ramp entering the mainline, whereas a diverge event is defined as a vehicle leaving its lane and entering the off-ramp lane. Similar to cut-in behavior, a merge/diverge event often causes a conflict or even a crash with the following vehicle. The corresponding minTTC was calculated as the conflict indicator.
- Intersection conflict event: This type of behavior is measured using minimum post encroachment time (minPET) (*52*) to describe the conflict severity between two vehicles that encroach on each other within a small spatiotemporal window.

## Freeway/Expressway Cut-in/Merge/Diverge

It has been demonstrated that CitySim has significantly more cut-in, merge, and diverge events than HighD and NGSIM. Two example locations from CitySim—Expressway A and Freeway C—were compared with HighD and NGSIM (US-101 and I-80). The NGSIM

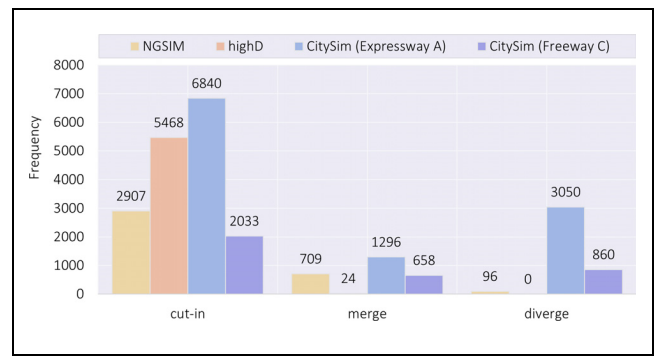

**Figure 11.** Safety-critical events extracted from each dataset.

Lankershim Boulevard Dataset was not included because it contains an urban scene rather than a freeway/expressway. Figure 11 shows that two locations from CitySim had 8,873 cut-in events, whereas HighD had 5,468, and NGSIM only 2,907 cut-in events. The two CitySim locations had 1,954 merge events and 3,910 diverge events. In contrast, the merge and diverge sample sizes of HighD and NGSIM were very small. Given that cut-in, merge, and diverge events often cause severe conflicts or even crashes, they are typically heavily tested for autonomous vehicle safety. As shown in Figure 11, CitySim provided these types of events to facilitate thorough safety research and development.

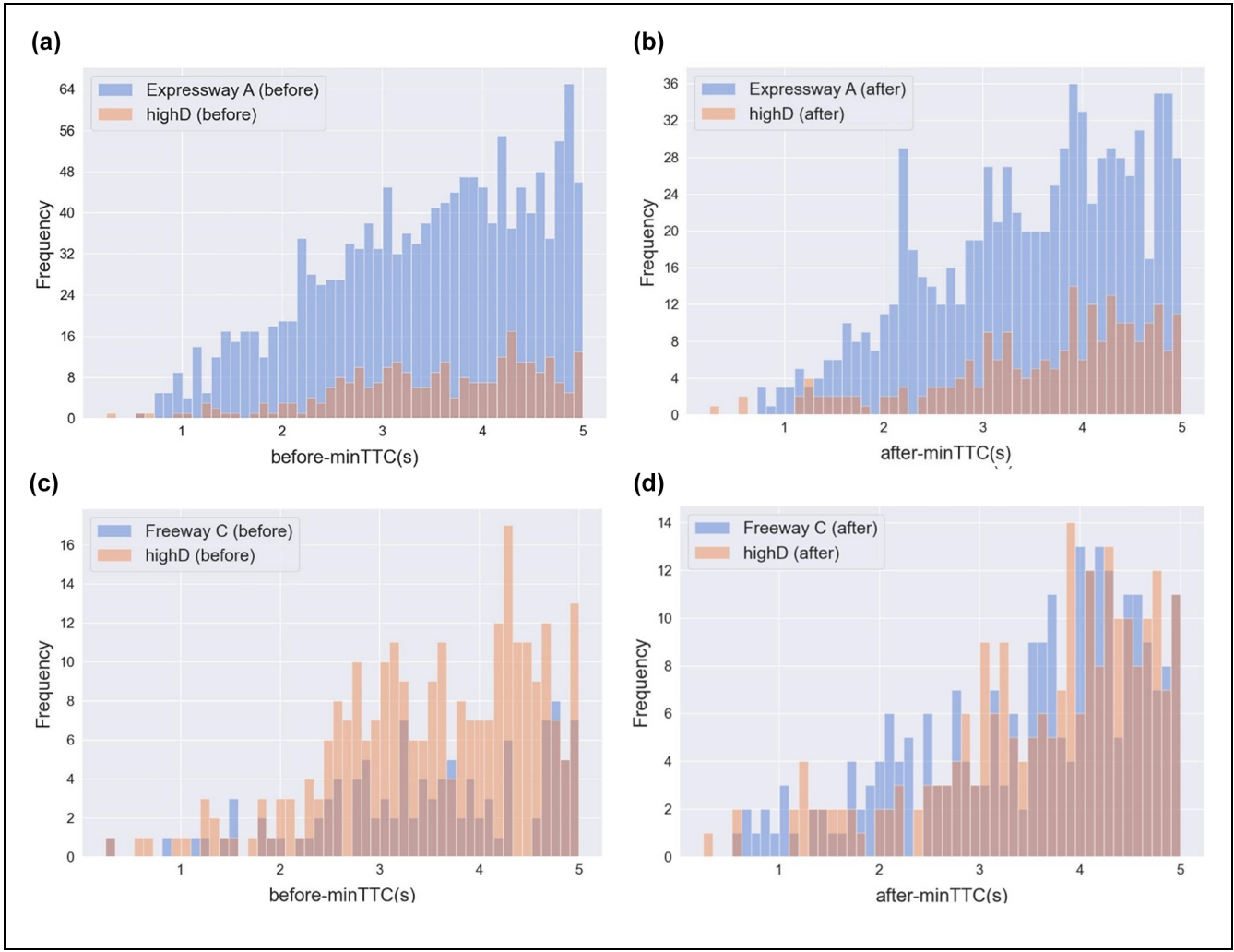

**Figure 12.** Cut-in-related conflict (minTTC < 5.0 s) comparison between CitySim and HighD: (*a*) CitySim Expressway A versus HighD (before) (*b*) CitySim Expressway A versus HighD (after) (*c*) CitySim Freeway C versus HighD (before), and (*d*) CitySim Freeway C versus HighD (after).

In addition to the larger sample sizes of safety-critical events compared with other datasets, the conflicts in CitySim were demonstrably more severe. Severe safety-critical events are defined as events with a minTTC/minPET less than a given threshold. The lower the conflict threshold, the more likely it is to cause a crash. In this comparison, a threshold of 5.0 s was adopted to show the distribution of small minTTC/minPET; basically, events with small minTTCs/minPETs are more meaningful for safety research. Figure 12*a* shows that, compared with HighD, Expressway A had more cut-in events with a before-minTTC and after-minTTC less than 5.0 s. In addition, Figure 12*b* demonstrates that Freeway C also had more severe cut-in events involving the following vehicle in the target lane. With regard to merge events, both Expressway A and Freeway C had more severe merge events compared with HighD (Figure 13). NGSIM was not considered in this comparison owing to its reported trajectory errors, which were found to generate abnormal minTTCs. Figure 14 illustrates some of the safety-critical events observed in CitySim.

## Intersection Conflicts

Robust safety conflict calculations require accurate vehicle geometric representation. Center-point-based conflict analysis can fail to detect certain safety-critical events or undermine their severity. As shown in Figure 15, a larger number of critical conflict events were identified when using bounding-box-based measurement compared with center-point-based measurements. Furthermore, the conflict events were more severe than those identified by center-point trajectories. As demonstrated in Figures 16 and 17, the distribution of the critical conflict events at the intersections were significantly different. This demonstrates the significant bias error introduced by the center-point-based vehicle trajectories. Therefore, the rotated bounding box information, as provided by CitySim, is believed to have the potential to benefit safety evaluation much more than several other datasets.

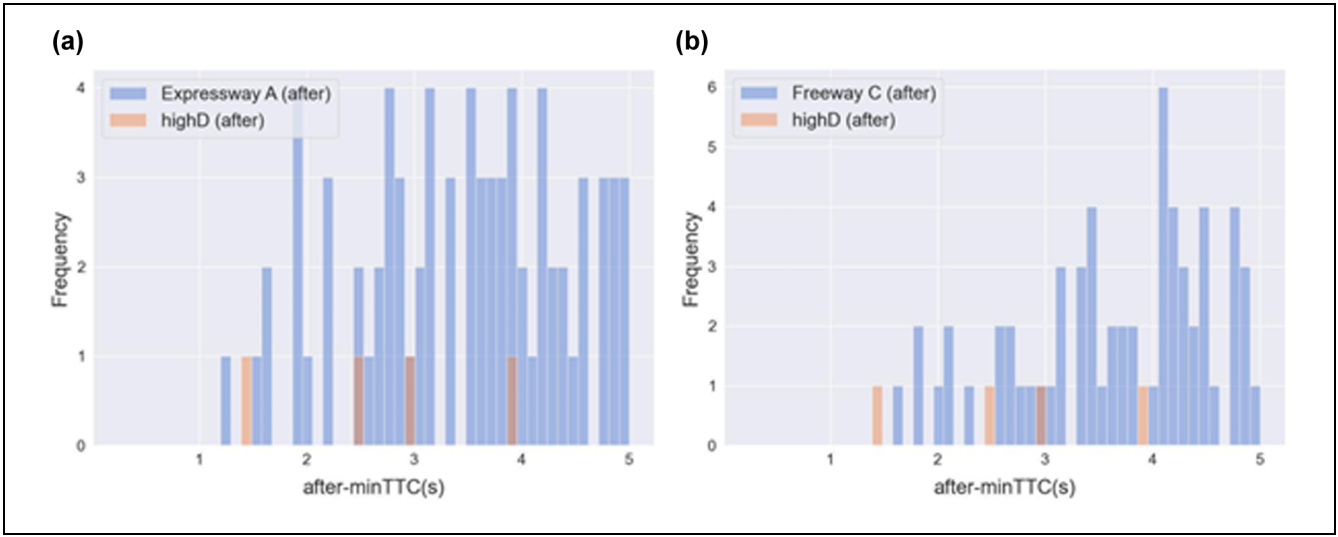

**Figure 13.** Merge-related conflict (minTTC < 5.0 s) comparison between CitySim and HighD: (*a*) CitySim Expressway A versus HighD (after) and (*b*) CitySim Freeway C versus HighD (after).

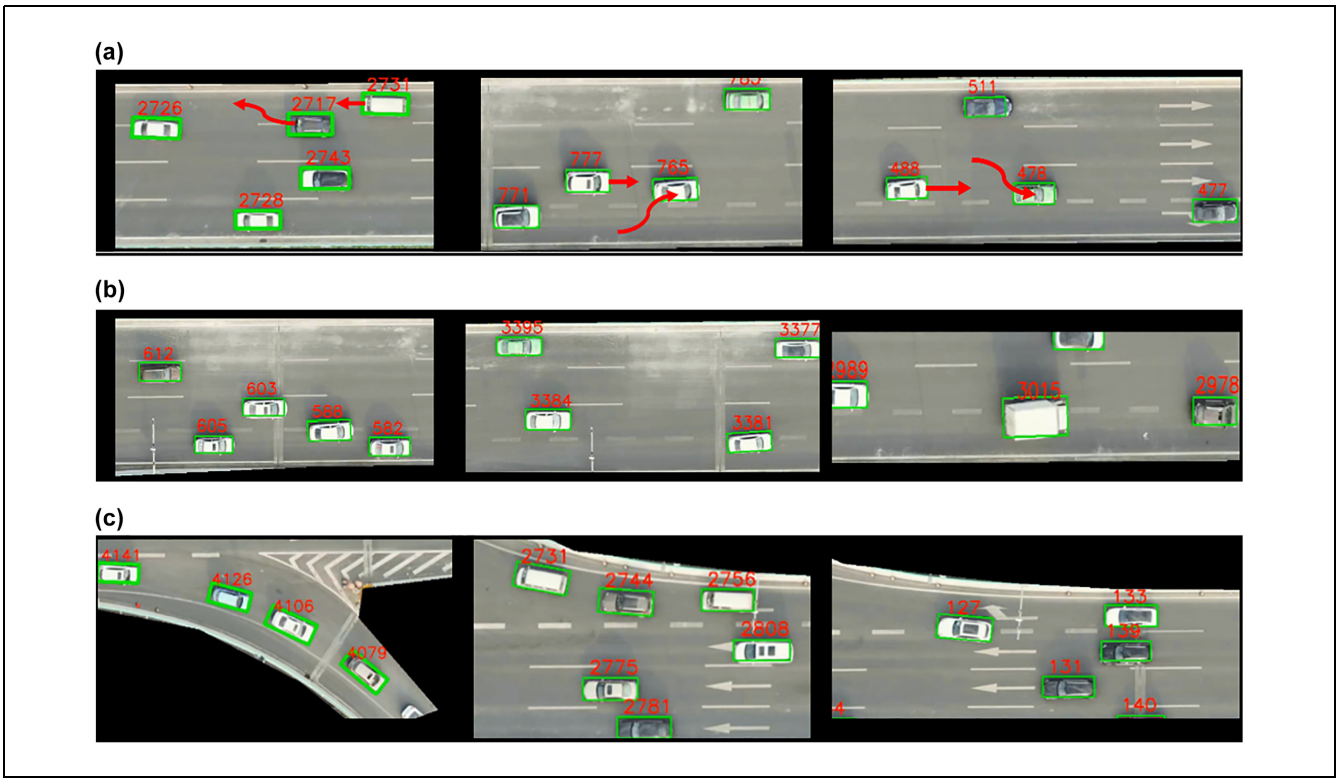

**Figure 14.** Safety-critical events extracted from CitySim: (*a*) cut-in events, (*b*) merge events, and (*c*) diverge events.

## Toward Digital Twins

Traffic digital-twin research has been growing in popularity over the past few years. As stated, a digital twin is a digital replica of all the traffic participants. It enables

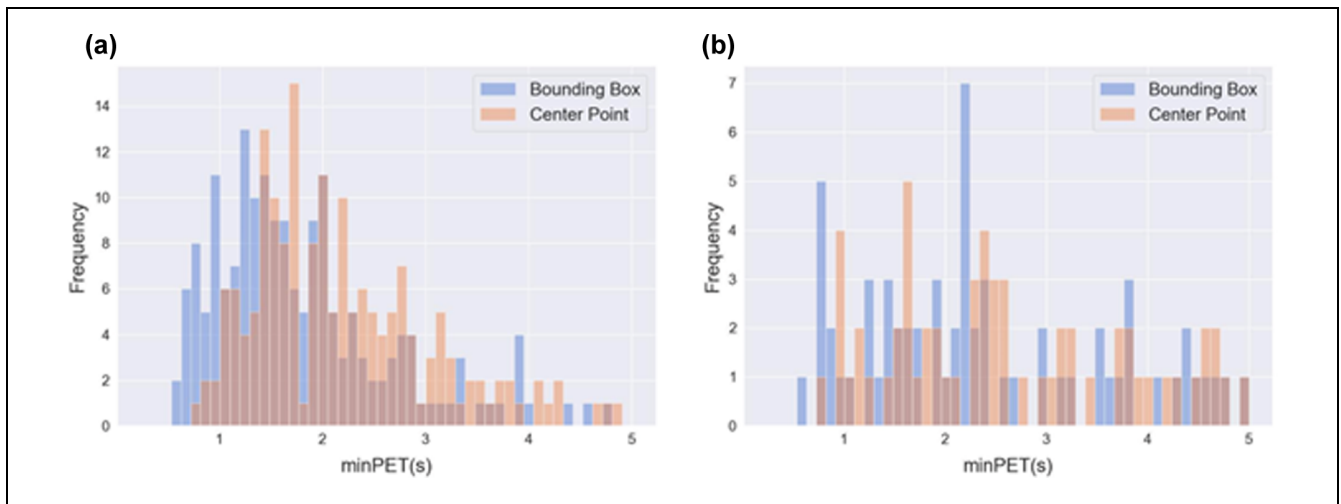

**Figure 15.** Critical conflict events at two CitySim intersections measured by different minPET thresholds: (*a*) Intersection A and (*b*) Intersection G.

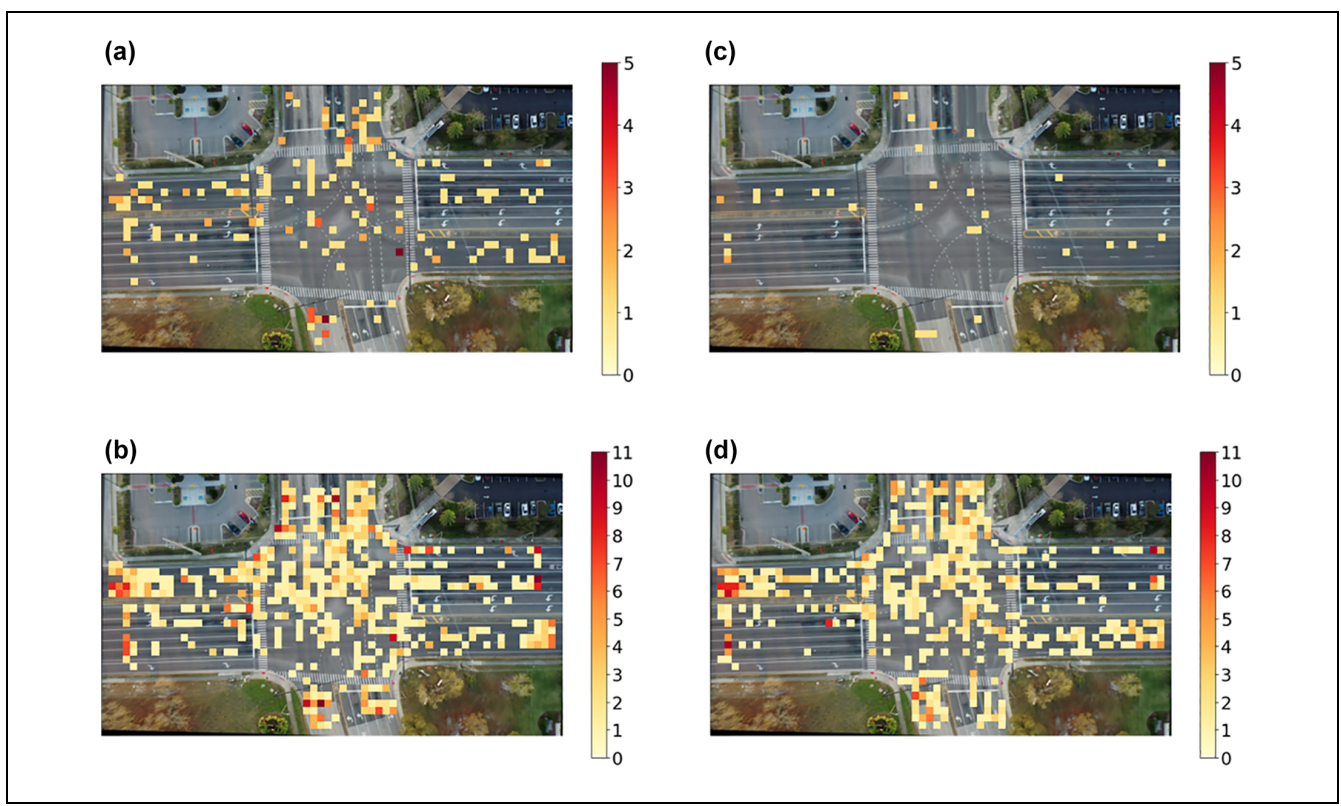

**Figure 16.** Heatmap of critical conflict event distribution at Intersection A measured by different minPET thresholds: (*a*) bounding box minPET < 1.0 s (204); (*b*) bounding box minPET < 2.5 s (994); (*c*) center point minPET < 1.0 s (36); (*d*) center point minPET < 2.5 s (797).

*Note:* The numbers in parentheses indicate the number of events.

real-time monitoring and synchronization of multiple real-world vehicles via their digital replicas. By projecting and simulating the traffic participants (vehicles, pedestrians, signal lights, etc.) in the virtual environment, the real-world traffic condition is reflected in a digital world. This allows the traffic flow, traffic safety, and mobility to be monitored and optimized in real time. Furthermore, it is effective for research and field implementation of connected and autonomous vehicles, as it can be simulated in a high-fidelity virtual world including vehicle dynamics, weather conditions, environmental perceptions, and other in-the-loop factors, which cannot be achieved in traditional microscopic traffic simulation.

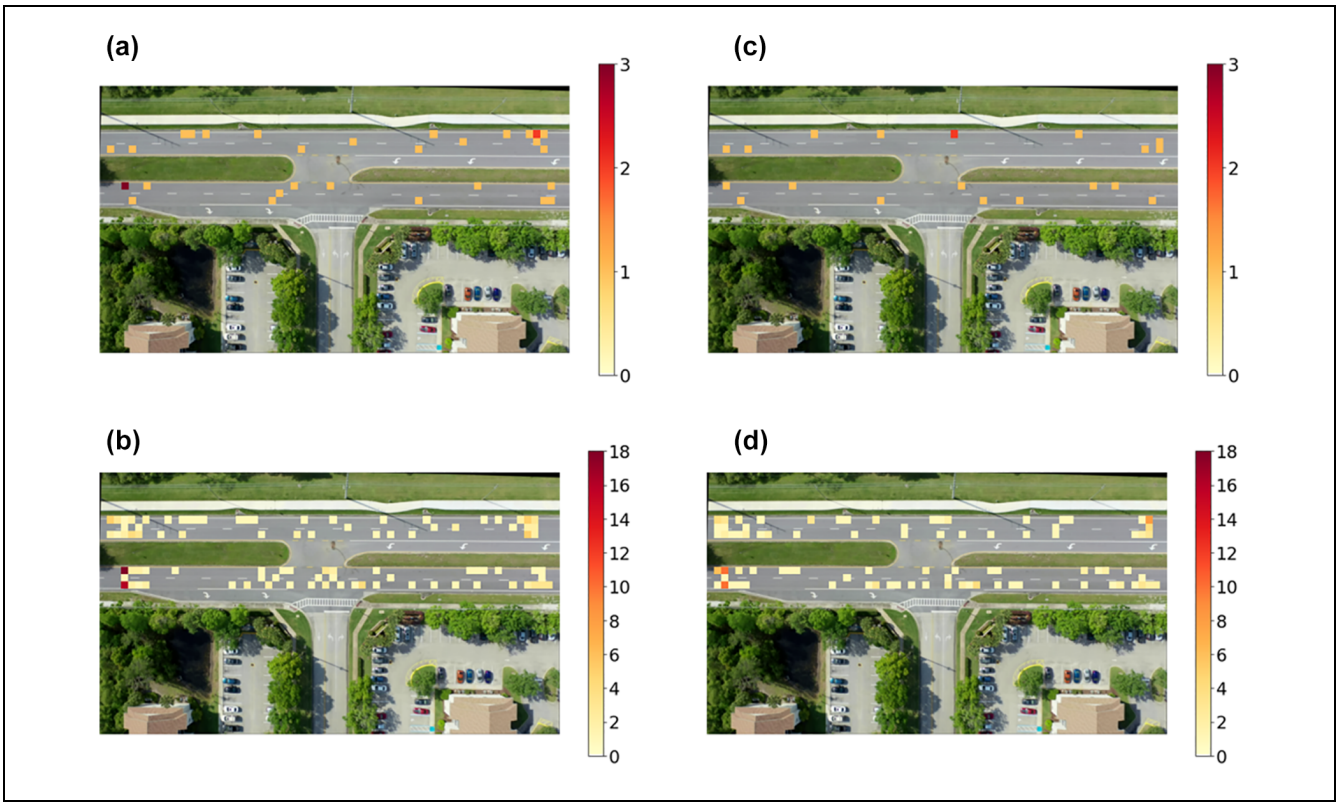

**Figure 17.** Heatmap of critical conflict event distribution at Intersection G measured by different minPET thresholds: (*a*) bounding box minPET < 1.0 s (32); (*b*) bounding box minPET < 2.5 s (177); (*c*) center point minPET < 1.0 s (20); (*d*) center point minPET < 2.5 s (152). *Note*: The numbers in parentheses indicate the number of events.

To facilitate digital-twin-based applications, CitySim provides a 3D base map for each location. The 3D model was built based on geographic information system (GIS) data (e.g., aerial image, elevation, and point cloud data), road vector data (OpenStreetMap), and Google Streetview. The professional road design tool RoadRunner was used to create the 3D simulation model. Aerial image, point cloud, and road vector data were used to draw the road network. Each lane with its marks and lines was duplicated in a way that matched the GIS data to ensure accuracy. The buildings, roadside signs, and plants were built by architects using Google Streetview as a reference. Figure 18 illustrates a few of the completed 3D base map snapshots of the CitySim locations.

In addition to base maps, CitySim also provides signal timing data related to signalized intersections. Each record in the signal data CSV file represents a change in one of the signals. The data are represented by eight different digits that correspond to different signal phases at the intersection.

Figure 19 illustrates a general framework for using CitySim assets to create digital-twin-based applications. CitySim provides accurate vehicle trajectories from its drone data, which can be used to calibrate microscopic traffic patterns. Then, the calibrated traffic patterns and 3D base maps can be utilized in a cosimulation platform that integrates microscopic simulation (e.g., SUMO) and driving simulation (e.g., Carla). By simulating the virtual testing environment, vehicle dynamics, and -sensors, a human-in-the-loop simulator experiment can be conducted using a cosimulation platform, which fulfills a digital-twin-based application that connects the virtual and physical worlds.

## Conclusions

This paper introduces CitySim: a new birds-eye-view video-based trajectory dataset that particularly aims to facilitate safety research. CitySim comprises vehicle interaction trajectories extracted from 1,140-min of video recordings, and covers a variety of locations, including freeway basic segments, freeway weaving segments, expressway segments, signalized intersections, stop-controlled intersections, and control-free intersections. CitySim was generated through a five-step procedure that integrates both advanced computer vision technology and manual error-checking, thereby ensuring high accuracy. The advantage of CitySim over other datasets is that it contains more safety-critical events. The safety events can be analyzed in a more accurate way given that vehicle rotated bounding box information is provided. These advantages provide sufficient support use in safety-oriented research. In addition, CitySim provides digital-twin features including 3D base maps and signal timings, which enable a more comprehensive testing environment for safety research, such as that required to ensure autonomous vehicle safety. In the future, more locations that contain intensive vehicle interactions will be included, such as weaving segments, unsignalized intersections, and mid-blocks; accordingly, a wider range of conflict types will be captured. Corresponding 3D base maps will also be created. To facilitate traffic simulation, traffic flow parameters will be provided and typical traffic flow models calibrated. In addition, a data visualization tool will be developed so that users can better screen and visualize the vehicle trajectories.

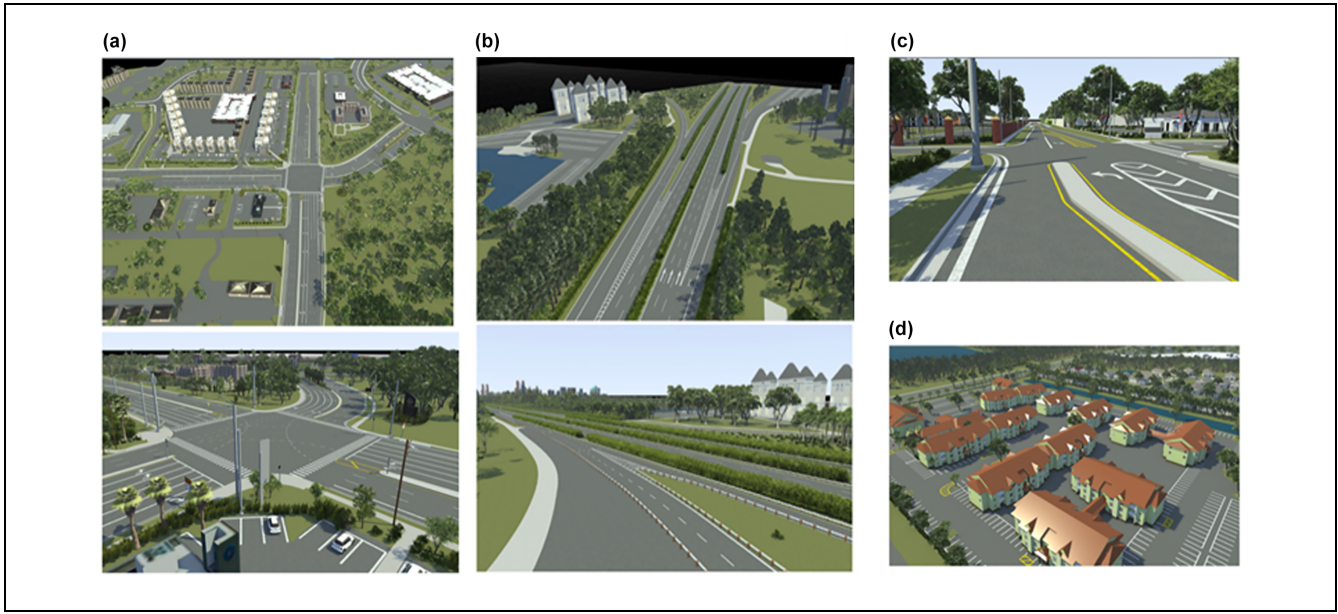

**Figure 18.** Three-dimensional (3D) models for each of CitySim's location: (*a*) intersection 3D map example, (*b*) highway 3D map example, (*c*) landmark detail, and (*d*) building detail.

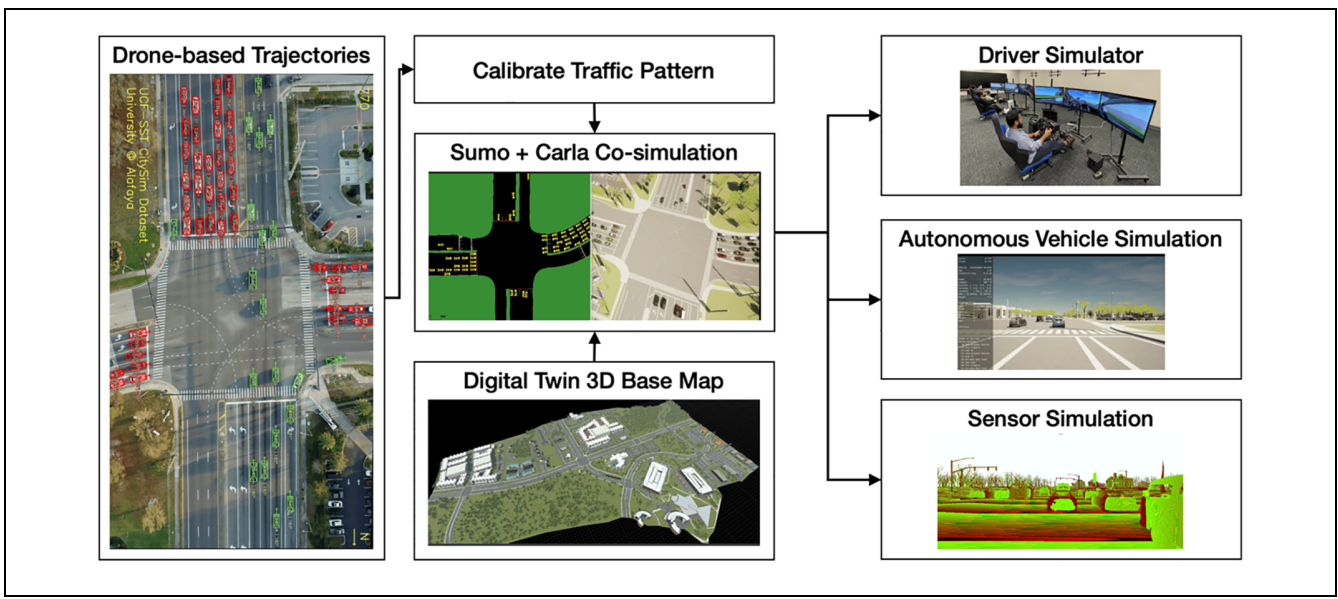

**Figure 19.** Digital-twin-based simulation framework that integrates both virtual and physical worlds.

## Author Contributions

The authors confirm contribution to the paper as follows: study conception and design: O. Zheng, M. Abdel-Aty; data collection: O. Zheng, L. Yue, Z. Wang, N. Mahmoud; analysis and interpretation of results: M. Abdel-Aty, L. Yue, O. Zheng, A. Abdelraouf; draft manuscript preparation: M. Abdel-Aty, L. Yue, A. Abdelraouf, O. Zheng. All authors reviewed the results and approved the final version of the manuscript.

## Declaration of Conflicting Interests

The authors declared no potential conflicts of interest with respect to the research, authorship, and/or publication of this article.

## Funding

The authors received no financial support for the research, authorship, and/or publication of this article.

## Data Accessibility Statement

The dataset is available online at https://github.com/ozheng1993/UCF-SST-CitySim-Dataset.

## ORCID iDs

Ou Zheng  https://orcid.org/0000-0001-6313-8566
Mohamed Abdel-Aty  https://orcid.org/0000-0002-4838-1573
Lishengsa Yue  https://orcid.org/0000-0002-0864-0075
Amr Abdelraouf  https://orcid.org/0000-0001-9068-6664
Zijin Wang  https://orcid.org/0000-0002-3285-433X
Nada Mahmoud  https://orcid.org/0000-0002-5424-4591